\documentclass[runningheads]{llncs}
\usepackage{graphicx}
\usepackage{amsmath,amssymb} 
\usepackage{color}

\begin{document}
\title{DeepTracking-Net: 3D Tracking with Unsupervised Learning of Continuous Flow} 

\titlerunning{DeepTracking-Net}
%
\author{Shuaihang Yuan
\and
Xiang Li
\and
Yi Fang\thanks{indicates corresponding author.}
}
%
\authorrunning{S. Yuan, X. Li and Y. Fang}
%

\institute{NYU Multimedia and Visual Computing Lab, USA \and
New York University Abu Dhabi, UAE\and
New York University, USA\\
\email{\{sy2366, xl1845, yfang\}@nyu.edu}}
\maketitle              

\begin{abstract}

This paper deals with the problem of 3D tracking, i.e., to find dense correspondences in a sequence of time-varying 3D shapes. Despite deep learning approaches have achieved promising performance for pairwise dense 3D shapes matching, it is a great challenge to generalize those approaches for the tracking of 3D time-varying geometries. In this paper, we aim at handling the problem of 3D tracking, which provides the tracking of the consecutive frames of 3D shapes. We propose a novel unsupervised 3D shape registration framework named DeepTracking-Net, which uses the deep neural networks (DNNs) as auxiliary functions to produce spatially and temporally continuous displacement fields for 3D tracking of objects in a temporal order. Our key novelty is that we present a novel temporal-aware correspondence descriptor (TCD) that captures spatio-temporal essence from consecutive 3D point cloud frames. Specifically, our DeepTracking-Net starts with optimizing a randomly initialized latent TCD. The TCD is then decoded to regress a continuous flow (i.e. a displacement vector field) which assigns a motion vector to every point of time-varying 3D shapes. Our DeepTracking-Net jointly optimizes TCDs and DNNs' weights towards the minimization of an unsupervised alignment loss. Experiments on both simulated and real data sets demonstrate that our unsupervised DeepTracking-Net outperforms the current supervised state-of-the-art method. In addition, we prepare a new synthetic 3D data, named SynMotions, to the 3D tracking and recognition community.

\end{abstract}

\section{Introduction}

\begin{figure}[ht!]
\centering
  \includegraphics[width=\textwidth]{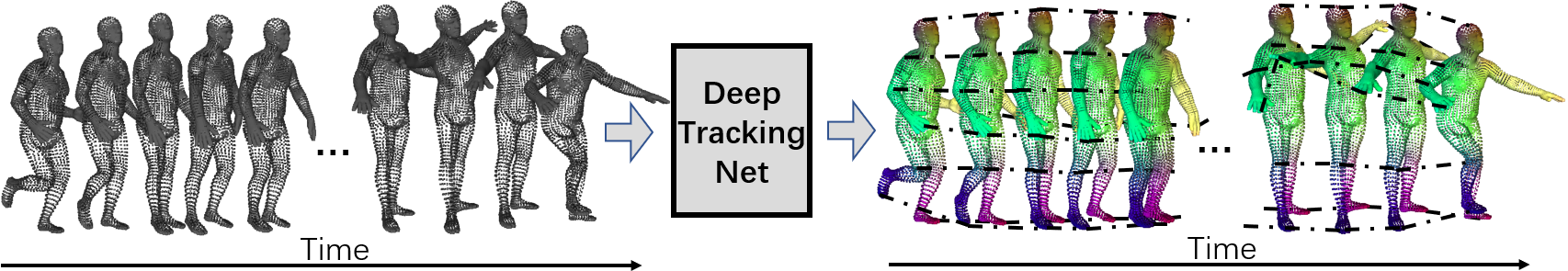}
  \caption{
Illustration of our proposed method. Given a set of time-varying 3D shapes, our DeepTracking-Net generates the spatially and temporally continuous flow for 3D tracking of time-varying shapes.}
  \label{fig:overview} 
\end{figure}
Given a sequence of time-varying 3D shapes, 3D tracking aims to find a temporally and spatially continuous flow which assigns a motion vector to every point on 3D shapes ,thus implicitly capturing dense correspondences over the temporal sequence of 3D shapes. It plays an important role in many real-world applications such as 3D animation, augmented reality \cite{chai2002three} and 3D rigging \cite{baran2007automatic,pan2009automatic}. 

As discussed in \cite{huang2016volumetric}, the standard process of 3D tracking starts with identifying the shape correspondence between each observed frame and a 3D template shape, and follows by deforming the 3D template towards the observed frames according to the shape correspondences. To improve the tracking performance, in addition to developing better 3D deformation models, prior efforts paid great attention to machine-learning-based techniques to discriminatively determine the shape correspondence. In \cite{varol17_surreal}, the supervised point signature is learned for 3D shape correspondence by training a neural network on the annotated data set. The learning-based point signature is used to guide the identification of the shape correspondence according to a certain type of similarity metric.

The great success of deep learning in various computer vision applications inspires the researchers to develop the unified paradigm that uses deep neural networks (DNNs) to deform the 3D template shape directly to the observed one without explicitly identify the shape correspondence \cite{wang20193dn,groueix20183d,niemeyer2019occupancy,mescheder2019occupancy}. In \cite{groueix20183d}, authors proposed shape deformation networks to jointly encode 3D shapes and their correspondences by factoring the 3D surface representation into (i) a template which parameterizes the 3D surface, and (ii) a learned global feature vector which parameterizes the geometric transformation of the template into the input surface. Their shape deformation networks demonstrate promising performance in shape correspondence, which motivates us to continue this line of research towards a deep learning-based 3D tracking in this paper. 

There are three major challenges for the problem of time-varying 3D shape tracking: 1) 3D tracking aims to find correspondences for time-varying 3D shapes in both space and time. The temporal information among consecutive frames of 3D shapes is critical in identifying shape correspondence in addition to 3D spatial information, 2) 3D tracking sometimes handles the large time-varying shape deformation which transforms a template shape to observed one quite intractable, and 3) the design of an explicit encoder to extract spatial-temporal feature from time-varying 3D unstructured point clouds is challenging as the discrete convolutions operator assumes the data formed with regular grid structure (i.e. 3D voxel and 2D image).

In this paper, to address those three technical issues, we propose a novel unsupervised 3D tracking approach, named DeepTracking-Net, which is based on the newly introduced latent feature representation named ``temporal correspondence descriptor (TCD)''. In addition to the 3D geometric information for each frame of 3D shape, the proposed TCD takes into account of temporal correspondences among consecutive frames. As shown in Figure \ref{fig:overview}, the CF-decoder takes input as TCD and regresses the globally continuous flow for consecutive frames. As you can see in Figure \ref{fig:overview}, the DeepTracking-Net jointly updates the TCD features and network parameters towards the minimization of an unsupervised global shape correspondence loss among a sequence of 3D shapes. Therefore, our DeepTracking-Net eliminates the hand-craft design in feature encoder but directly employs a learnable latent vector to encode both temporal and spatial essence for time-varying shape correspondence for 3D tracking. In addition, DeepTracking-Net does not use the 3D template as the bridge model to globally corresponding the consecutive 3D shapes; instead, it optimizes a global TCD that is decoded to spatially and temporally continuous flow for 3D tracking.

\begin{figure*}[ht!]
  \includegraphics[width=\textwidth]{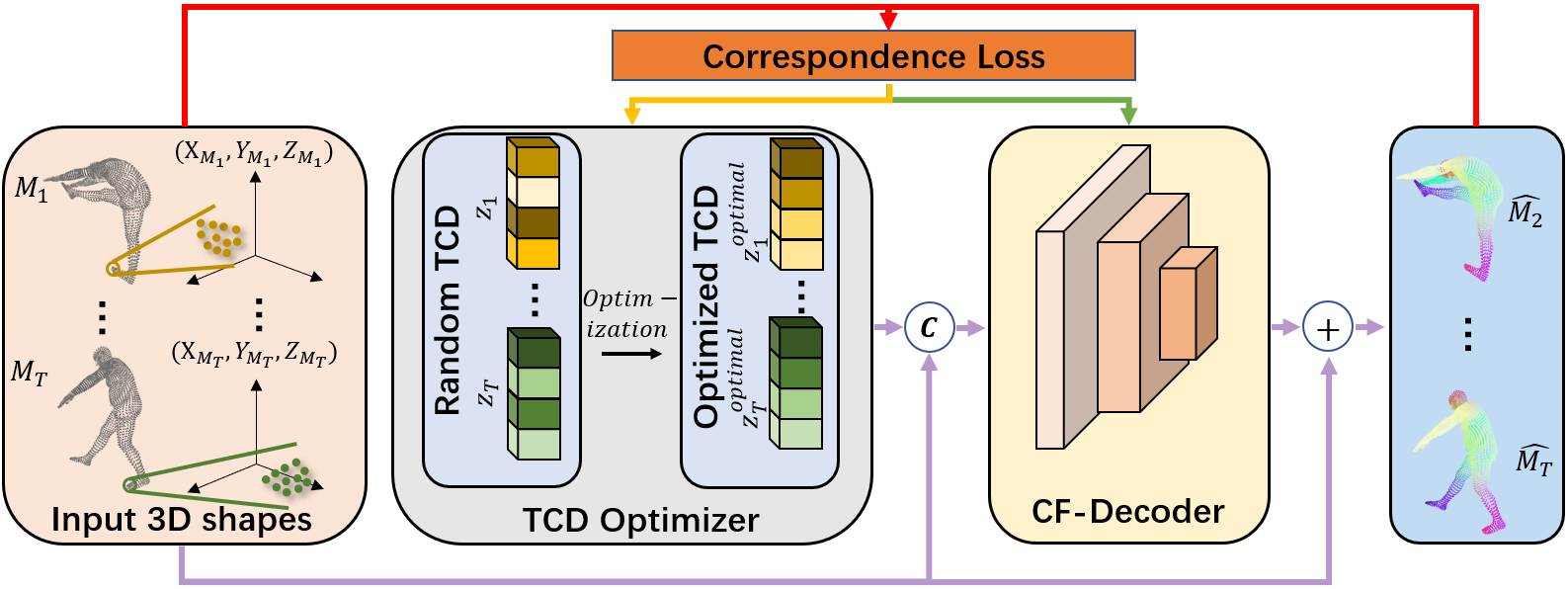}
  \caption{
  The pipeline of DeepTracking-Net. It takes a sequence of 3D shapes $M_{1},M_{2}, ... M_{T-1}$ as inputs and produces the flow $\hat{M_{2}},\hat{M_{3}}, ... \hat{M_{T}}$ from one frame to the next. In the figure, the purple route indicates the forward path and the red route is used to present the computation of the unsupervised loss. Moreover, the green and yellow routes represent back-propagation which is used to update the TCD and CF-Decoder respectively. The process of updating CF-Decoder is only activated in the training stage and it is depreciated during the inference period.
  }
  \label{fig:pipeline} 
\end{figure*}

The overview of our method is presented in Figure \ref{fig:pipeline}. We summarize our contributions as follows:

\begin{itemize}

\item We design a novel learnable latent vector, named temporal correspondence descriptor (TCD), to encode both temporal and spatial essence for time-varying shape correspondence for 3D tracking. In this way, we avoid the design challenges of the explicit encoder for feature learning from unstructured 3D point clouds.

\item We develop a template-free 3D tracking approach which addresses the challenges of large shape deformation for time-varying 3D shapes in consecutive frames.

\item We introduce the DeepTracking-Net, a novel unsupervised 3D tracking paradigm that can even outperform the current supervised state-of-the-art method based on test results on D-Faust motion dataset.

\item We propose a new synthetic motion data set that contains dense ground-truth correspondences among each frame. The synthetic data set will be publicly available.
\end{itemize}

\section{Relate Work}
In this section, we divided our related work into two main fields: 1) 3D shape tracking. 2) descriptor learning.
\subsection{3D Shape Tracking}
Previous methods on finding shape correspondences assume a common model structure that shares among a set of data and the corresponding points are identified by minimizing the metric such as geodesic distances. The performance of these approaches is highly depending on the initial starting point since it is a non-convex function \cite{bronstein2006efficient}.
Hence, many point signatures such as HKS \cite{sun2009concise} are used to address this problem by finding the most similar point signatures. With the fast development of deep learning, many approaches are proposed to bring better ways for the 3D point signature and shape descriptor learning. Voxel grid is the most regular data representation in a 3D space and it discretizes the 3D space into small volumes. Since the 3D voxel is regular and ordered, we can easily generalize the learning methods in 2D to 3D field. Wu et al. \cite{VOXel-deep} first extend the learning method of CNN from 2D into 3D voxel. Then, \cite{VOXel-deep,VOx-3DConv} are proposed to obtain shape descriptors from voxel grid data. However, volumetric data representations are of great difficulty to capture shape details and smooth shape surfaces \cite{wu20153d,choy20163d}. In addition to the voxel grid data representation, the point cloud data is the most common data to be obtained. However, due to the unorder and irregular properties of a point cloud, it is not practical to directly apply CNN on point cloud data. Qi et al. \cite{qi2017pointnet} propose PointNet to obtain a global shape descriptor. Many following works using the PointNet as the encoder to perform the shape feature extraction tasks \cite{achlioptas2017learning,yang2017foldingnet}. The mesh data is another 3D representation composed of a set of points, edges, and faces. Different from the point cloud, mesh data only contains points on the surface of the object. Since meshes are composed of vertices and edges, they can be viewed as graphs. FeastNet \cite{verma2018feastnet} propose a graph convolution to analyse 3D mesh shapes and DCN \cite{xu2017directionally} introduce a directionally convolutional network to extract 3D shape features. Those deep learning-based methods are proposed to analyze 3D shapes and future improve the matching result as they obtain more accurate feature by neural networks.

\subsection{Descriptor Learning}
As the development of deep learning, a mountain of learning techniques is proposed. Recent researchers focus on techniques that can explore the features of data automatically. GANs, Generative Adversial Networks, is first proposed by Goldman et al. \cite{goodfellow2014generative}. This technique automatically embeds the input data by training a discriminator and generator. The generator aims to generate the target shape and discriminator aims to discriminate whether the generated shape is acceptable. In addition to the GANs, the auto-encoder is another technique that aims to produce the output exactly the same as the input to obtain the bottleneck information. Many recent works \cite{stutz2018learning,groueix2018atlasnet,wu2018learning} design learning algorithms based on the auto-encoder architecture. Different from the auto-encoder, the encoder-free network is also proposed to use the latent vector as a bottleneck feature. In \cite{tan1995reducing}, Tan et al. uses the back-propagation to jointly optimize the latent space vector and the decoder network. Base on this idea, researchers propose approaches \cite{reddy1998input,qunxiong2006dimensionality,park2019deepsdf} to address different tasks such as shape completions and noise reduction.

\section{Approach}
\subsection{Problem Statement}
In this section, we first define the tracking problem as the following. Given a sequence of 3D shapes $\mathbf{M}=(M_i)_{i=1}^{T}$ contains $T$ shapes which are observed over an ordered time sequence. Each shape $M_i$ is represented by a $N_i\times3$ coordinate matrix, where $N_i$ denotes the number of observed points in the $i^{th}$ frame. 3D object tracking aims at estimating the dense correspondences $\mathcal{C}_{i}$ from one frame to the next frame, where this $\mathcal{C}_{i}$ can be formulated as a mapping function $\mathcal{C}_{i}(\cdot):(\mathbb{R}^3) \mapsto (\mathbb{R}^3)$. 

Instead of directly predicting the discrete dense correspondence for consecutive shapes, in this paper, we aim to use the deep neural network to learn a continuous flow function $\mathcal{D}_{\theta}(\cdot):(\mathbb{R}^3) \mapsto (\mathbb{R}^3)$ which estimates the flow from the one source 3D shape $M_i$ to the consecutive 3D shape $M_j$, where $\theta$ denotes network parameters. The problem is then formulated as finding the optimal parameter $\theta$ that minimizes alignment loss between transformed source shape and target shape, formulated as:
\begin{equation}
 \theta^{optimal}=\mathop{\arg\min}_{\theta} \sum_{ij} \mathcal{L}(\mathcal{D}_{\theta}(M_i), M_{j})
\end{equation}
where $\mathcal{L}(\cdot)$ is a loss function defined on the transformed source shape and target shape.

\subsection{Method Overview}\label{sc_overview}
In this section, we present our DeepTracking-Net for time-varying 3D shapes tracking. The pipeline of DeepTracking-Net is illustrated in Figure \ref{fig:pipeline}. Our method tracks 3D shapes by estimating a sequence of continuous flow. Our proposed method contains three components. The key is the temporal correspondence descriptors (TCD) which not only capture the 3D geometric features but also embed features from the temporal domain. The detail of this descriptor will be explained in section \ref{sc_tcd}. 

The second component is the continuous flow decoder (CF-Decoder) which decodes the TCD features to predict the flow from one frame to the next frame. We leverage the power of the deep neural network to build a continuous flow function. The details of this decoder network will be explained in section \ref{sc_decoder}. The third component is correspondence loss that measures the alignment between transformed source shape and the target shape, which will be illustrated in section \ref{sc_loss}. As shown in Figure \ref{fig:pipeline}, there are three routes presented in red, yellow and green lines. The red route represents the computation of our unsupervised loss. The yellow and green routes represent the back-propagation route which is used to update the TCD and continuous flow decoder respectively. We use the red route in both the training and the inference stage while the green route is only activated in the inference stage.

\subsection{TCD: Temporal Correspondence Descriptor}\label{sc_tcd}

\begin{figure}[ht!]
\centering
  \includegraphics[width=\textwidth]{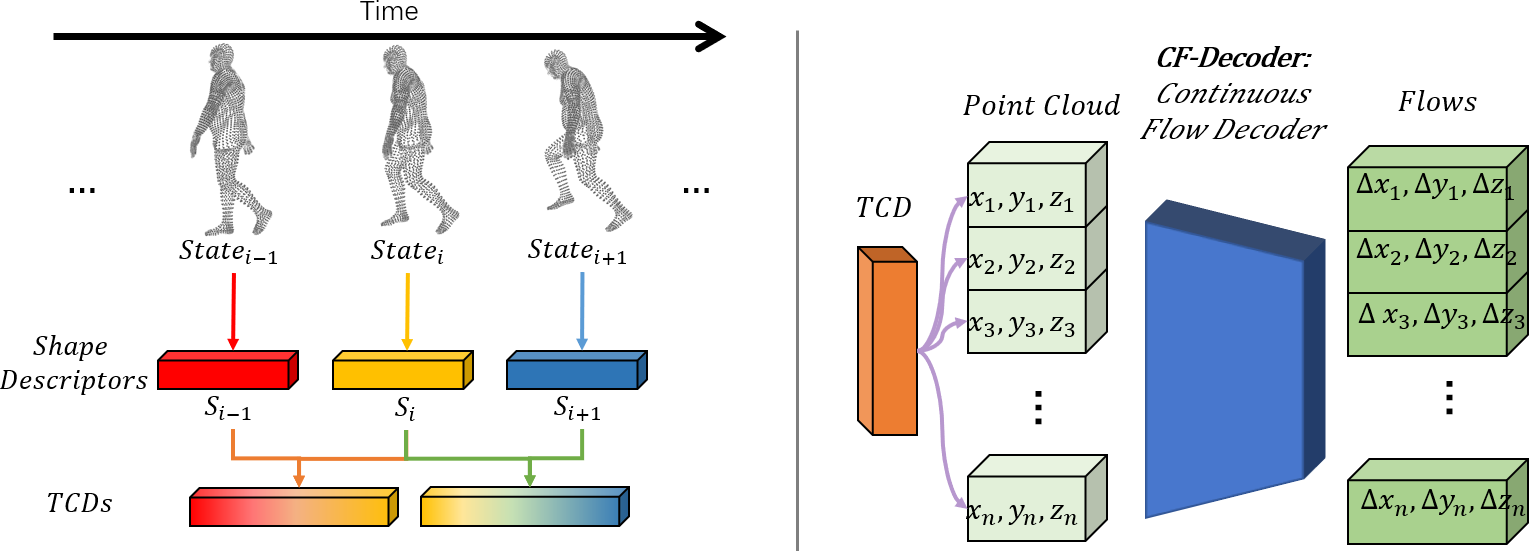}
  \caption{
\textbf{Left}: The illustration of TCD. For each state, we have a unique latent vector work as the shape descriptor and the target TCD is acquired by computing the Equation \ref{eq:z}. \textbf{Right}: The illustration of continuous flow decoder. The CF-Decoder predicts the displacement of each point. For a clear presentation, we only show one point cloud with its corresponding TCD. 
  }
  \label{fig:TCD_dec} 
\end{figure}

To track 3D objects in a sequence of consecutive frames, it is of great importance to capture both geometric features of each 3D shape as well as the temporal correlation among consecutive frames. To compute the geometric features, previous works usually need a feature encoder (i.e. PointNet \cite{qi2017pointnet}) to extract deep geometric features. However, the design of an appropriate feature encoder network is challenging due to the unstructured and unordered nature of 3D point clouds. More importantly, by conducting feature learning for each frame independently, the learned feature representations will only include the spatial information from each local frame, which can limit the tracking performance among a sequence of frames and as a whole. 
To address these two challenges, in this paper, we propose to leverage a novel learnable latent vector TCD as shown in the left part of Figure \ref{fig:TCD_dec} to encode both spatial and temporal essence for time-varying 3D shapes. By doing so, our method avoids the challenge from the explicit design of the feature encoding network and makes it easier for temporal information integration with the latent code.


For a given sequence of temporal 3D shapes $\mathbf{M}$, we set each 3D shape a unique state. We assume that the geometric contents of each state in $\mathbf{M}$ can be represented as a learnable latent vector $s_i$. Then, we formulate all $s_i$ as $\mathbf{S}=(s_i)_{i=1}^{T}\subset\mathbb{R}^{D_S}$, and initialize each $s_i$ from a Gaussian distribution, i.e., $s_i \sim \mathcal{N}(0,1)$. Given this representation, we further introduce the proposed temporal correspondence descriptor (TCD) that characterizes the transformation relationships from current state $i$ to next state. We denote the temporal correspondence descriptor as $\mathbf{Z}=(z_i)_{i=1}^{T}$, where $z_i \in \mathbb{R}^{D_{z}}$. The $i^{th}$ TCD is initially as follows:
\begin{equation}
    z_i = \left\{\begin{array}{lr}
    (1-\omega) z_{i-1}+ \omega s_i, & i >1 \\s_i, & i=1\end{array}\right.
    \label{eq:z}
\end{equation}
, where $\omega \in \mathbb{R}$ is a trainable weight parameter representing the influence from the previous frames. Notice that, from Equation \ref{eq:z}, each TCD feature $z_i$ is of the same dimension as $s_i$. 

By aggregating the weighted spatial features from previous states, our TCD representation naturally represents the spatio-temporal features lied in all previous states. The temporal weight parameters $\omega$ can be optimized during network training to determine the optimal values for temporal neighbors.

Although TCD does not compute correspondence explicitly, it encodes spatio-temporal essence from consecutive frames. The TCD is fed into continuous flow decoder to predict the continuous mapping (i.e.3D flow) from a set of consecutive frames to the next frame. The continuous mapping will then provide the dense correspondence based on the nearest neighbor matching among consecutive frames. For this reason, it is called a temporal correspondence descriptor (TCD).


\subsection{CF-Decoder: Continuous Flow Decoder}\label{sc_decoder}

  

To track the 3D shape deformation from one state to the next, the transformation process defined on $M_i \in \mathbf{M}$ can be formulated as:
\begin{equation}
    \mathcal{T}_{M_{i+1}}=M_i+\mathcal{D}(M_i)
    \label{eq:tracking}
\end{equation}
, where $\mathcal{D}:\mathbb{R}^{3}\mapsto \mathbb{R}^{3}$ is a flow function that predict the displacement of each point from $M_i$ to $M_{i+1}$. The task of tracking among temporal frames can be transformed to the problem of predicting the displacement of each point on $M_i$. Regarding the Motion Coherent Theory \cite{yuille1989mathematical}, it is of great significance to ensure the input and the continuous flow function $\mathcal{D}$ are continuous. To achieve this goal, we use the deep neural network to approximate the flow function as shown in the right part of Figure \ref{fig:TCD_dec}. We first pair each 3D shape $M_i$, which contains $N_i$ points, with its corresponding TCD $z_i$. For each point $x$ on shape $M_i$, we stack its coordinates with the $z_i$ to formulate the input $[z_i, x]$ and then we feed it into the decoder network to predict the point displacement. Here, we use a 7-layer Multilayer Perceptrons (MLPs) network with its parameters $\phi$ for the approximation of the flow function $\mathcal{D}$,
\begin{equation}
    \mathcal{D}(M_i)=MLP_{\phi}([M_i,z_i])
    \label{eq:dlt}
\end{equation}
, where the $[:,:]$ denotes pointwise concatenation. The displacement of each point $\mathcal{D}(M)$ can be estimated by the function $MLP_{\phi}:\mathbb{R}^{3+D_{z}}\mapsto \mathbb{R}^{3} $ and tracking result can be obtained by applying the Equation \ref{eq:tracking} on the calculated point displacements.

\subsection{Correspondence Loss}\label{sc_loss}
Our model is trained with a loss term $\mathcal{L}_{\phi,\mathcal{Z},\omega}$, where $\phi$ are all parameters in our CF-Decoder, $Z$ is the trainable TCD and the $\omega$ is the weight parameter. In this paper, we use Chamfer distance to measure the alignment between deformed shape $T_i$ and target shape $M_{i+1}$:

\begin{multline}
    \mathcal{L}_{\phi,\mathcal{Z},\omega}(\mathcal{T}_{M_{i+1}},M_{i+1})=\frac{1}{|\mathcal{T}_{M_{i+1}}|}\sum_{x\in \mathcal{T}_{M_{i+1}}}\min_{y \in M_{i+1}}\|x-y\|_{2}+ \\
    \frac{1}{|M_{i+1}|}\sum_{x\in M_{i+1}}\min_{y \in \mathcal{T}_{M_{i+1}}}\|x-y\|_{2} 
    \label{eq:loss}
\end{multline}
where $|\mathcal{T}_{M_{i+1}}|$ and $|M_{i+1}|$ denote the number of points on shape $\mathcal{T}_{M_{i+1}}$ and $M_{i+1}$ respectively. The Chamfer distance measures the two-way average distance between each point in transformed shape and its nearest point in the target shape, which guarantees the bilateral alignment both forward and backward. Note that Chamfer loss does not need ground truth correspondences for loss computation.

\subsection{Optimization Strategy}
In section \ref{sc_tcd}, we define a set of trainable latent vectors $\bold{S}$, one ($s_i \in \bold{S}$) for each frame and the temporal weight $\omega$. One should note that our loss functions defined above are computed in a pure ``unsupervised" way. During the training process, both $\bold{S}$ and $\omega$ are optimized along with the network parameters $\phi$ using a stochastic gradient descent-based algorithm. The final loss term $\mathcal{L}_{\phi,\mathcal{S},\omega}$ is the summation of all paired Chamfer distance between frame $i$ and frame $i+1$. During the training process, we aim to find the optimum $\phi$, $S$ and $\omega$ using back-propagation that minimize the loss function following:

\begin{equation}
    (\phi^{optimal}, S^{optimal}, \omega^{optimal})= \mathop{\arg\min}_{\phi,S,\omega} (\mathcal{L}_{\phi,\mathcal{S},\omega})
    \label{eq:train}
\end{equation}

After training, the learned decoder network can be regarded as prior knowledge of how to decode a source shape and the corresponding TCD to the continuous flow towards the shape in the next frame. In the testing phase, given a set of consecutive frames, we first determine the optimized spatio-temporal features (TCD) while fixing the parameters of the trained decoder network, $\phi^{optimal}$ and $\omega^{optimal}$. The TCD optimization process can be formulated as:
\begin{equation}
    \hat{Z'}= \mathop{\arg\min}_{Z'} (\mathcal{Z'},\mathcal{L}_{\phi^{optimal},\omega^{optimal}})
    \label{eq:test}
\end{equation}
After optimization, the optimal TCD feature vector is fed into the decoder to predict a continuous mapping from current frames to the next frame. Finally, the tracking result can be acquired with the optimal $\hat{Z'}$ using Eq. \ref{eq:tracking} and Eq. \ref{eq:dlt}. The final dense correspondence can be estimated by nearest-neighbor matching between each two consecutive shapes.

\section{Experiment}
In this section, we conduct several experiments to validate the effectiveness of our proposed method for time-varying 3D tracking. In section 4.1, we first introduce the implementation details. In section 4.2, we present our newly introduced dataset SynMotions, which will be publicly available, as well as our evaluation metric used for quantitative evaluation. In section 4.3, we validate the performance of the proposed method. Then, in section 4.4, we compare our proposed method with recent state-of-the-art OFlow \cite{niemeyer2019occupancy} on D-Faust \cite{dfaust:CVPR:2017} dataset. In section 4.5, we perform experiments on the temporally inconsistent data that contains noise and partial observations to further validate the robustness of our model. Finally, we perform the ablation study to exam the effectiveness of our proposed TCD.

\subsection{Implementation Details}
Our proposed DeepTracking-Net is configured with the architecture presented in Figure \ref{fig:pipeline}. The continuous flow decoder network is composed of MLPs with seven hidden layers of size 256, 512, 1024, 2048, 512, 128, 3. For different experiments, the $\mathbf{Z}$ is set with different dimensions. We use an unsupervised loss as described in section 3.5 to train our neural networks. This loss term is used to update weight parameters $\omega$ and TCDs. Our model is trained with Adam optimizer with a learning rate of 0.001 and a batch size of 32, and it is implemented using PyTorch library and runs on an NVIDAI GTX 1080Ti GPU.

\begin{figure}[ht!]
\centering
  \includegraphics[width=\textwidth,height=6cm]{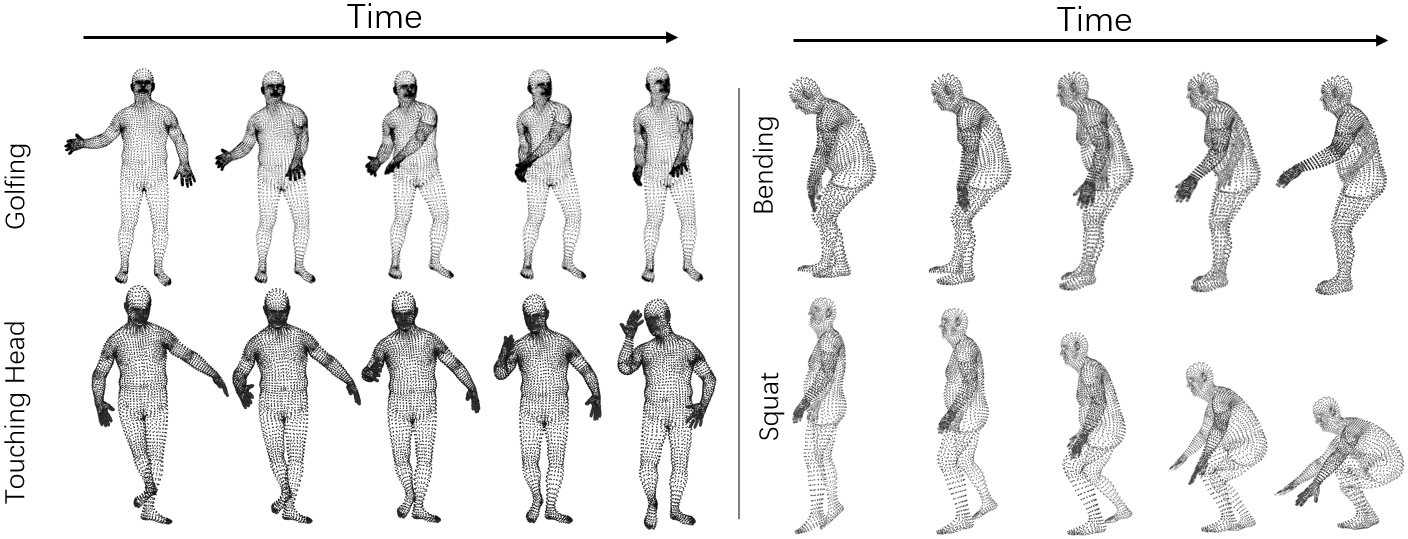}
  \caption{
    Selected example motions in our proposed SynMotion data set. 
  }
  \label{fig:data} 
\end{figure}

\subsection{Datasets and Evaluation Metric}

\noindent \textbf{Datasets.} One of the main reasons that motivate us to build our own synthetic data set is because it is not easy to obtain temporally clean point clouds in the real world. In addition, the existing motion datasets are often acquired with limited shape variations. Therefore, in this paper, we develop a synthetic train-test split dataset, named SynMotions. Selected example motions are shown in Figure \ref{fig:data}. We use SMPL human body model \cite{SMPL:2015} to generate temporal data. SMPL contains multiple human shapes but all shape models are generalized to contain 23 joints points and each joint point is controlled by 3 parameters. Therefore, there are in total of $69$ parameters to control the position of 23 joints. In addition to 23 joints, SMPL also uses 3 extra parameters to represent the root orientations. $69+3=72$ parameters in total are used to generate a human pose. However, those 72 parameters are unlimited. If we apply a random value for each parameter, it is most likely to generate an unrealistic human pose. To address these issues, we adopt the pose parameters presented in SURREAL \cite{varol17_surreal} to get random poses. To generate the temporal 3D shapes in a sequence of frames, we randomly select poses from the SURREAL pose database and perform interpolation along the time dimension.

In addition to our proposed SynMotions, we also use the Dynamic FAUST (D-Faust) data set \cite{dfaust:CVPR:2017} to evaluate our proposed method. This data set has various motions such as ``one leg jump'', ``shape arms'', and ``running on the spot'',  which are scanned from 10 real humans. In our experiment, we follow the same train-test split method as presented in \cite{niemeyer2019occupancy}, and then divide the D-Faust data set into a training set that has 105 motion sequences and a testing set that contains 9 sequences.

\begin{figure}[ht!]
\centering
  \includegraphics[width=\textwidth,height=5.5cm]{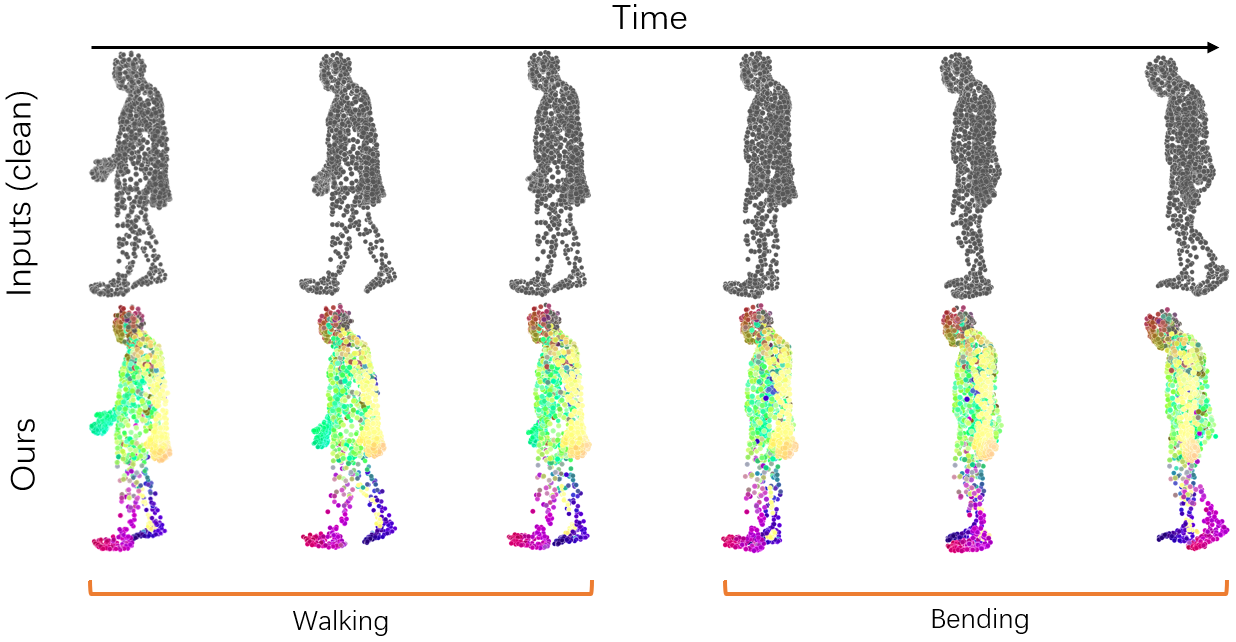}
  \caption{
  Visualized results on two motions: walking and bending. The consistent color-map over time-varying 3D shapes indicates the robust 3D tracking performance by our method.
 }
  \label{fig:cons_vis} 
\end{figure}

\noindent \textbf{Evaluation Metric.} Considering the experiment and evaluation metric used in Huang et al. \cite{huang2016volumetric} and Oflow\cite{niemeyer2019occupancy}, we adopt three evaluation metrics to evaluate the performance of our method. We first use the matching accuracy which is calculated by counting the ratio of accurate matching points within a threshold of Euclidean distance. If a point falls into a certain range of the corresponding point, it will be marked as the correct matching. Moreover, we use Chamfer distance between the transformed shapes and ground truth ones to evaluate the tracking performance.

\subsection{3D Tracking on Clean Data}\label{sc_clean}

\noindent\textbf{Experiment Settings:} In this experiment, we test our proposed method for 3D tracking on a clean dataset without using ground-truth correspondences for model training. Our model is firstly trained on the training split to learn the weight parameters of CF-Decoder $\mathcal{D}$ and temporal weight $\omega$. Then, we evaluate our model on unseen data. 

\noindent\textbf{Result Analysis: } 

\begin{table}[!htp]
\centering
\begin{tabular}{cccccccc}
\hline\hline
\multicolumn{8}{c}{SynMotions} 
\\\hline
\multicolumn{4}{c|}{Method} &\multicolumn{2}{c|}{Chamfer} &\multicolumn{2}{c}{Correspond.}
\\\hline

\multicolumn{4}{l|}{Ours, z=16} &\multicolumn{2}{c|}{0.018} &\multicolumn{2}{c}{0.075}
\\

\multicolumn{4}{l|}{Ours, z=64} 
&\multicolumn{2}{c|}{0.018} &\multicolumn{2}{c}{0.074}
\\

\multicolumn{4}{l|}{Ours, z=128} &\multicolumn{2}{c|}{0.016} &\multicolumn{2}{c}{0.071}
\\

\hline\hline
\end{tabular}
\label{tab:clean}
\caption{
Results on clean SynMotions clean dataset. The table shows the performance on the clean dataset with different sizes of $z$. We report Chamfer distance (lower is better) and correspond $\ell_2$-distance (lower is better). 
}
\end{table}

After trained on the training set of our SynMotions dataset, the temporal weight $\omega$ converges to $0.087$. We test the tracking performance on the test split of our proposed SynMotions data. The results are listed in Table 1 which show the tracking performance with different sizes of $z$ . From this table, one can see that the performance of our model is quite stable for the dimension of TCD. In addition, the visualized results are presented in Figure \ref{fig:cons_vis}.

\subsection{Comparison to Existing Work}
\noindent\textbf{Experiment Settings:} 
To validate the efficiency and effectiveness of our module, we compare the performance of our model with OFlow\cite{niemeyer2019occupancy} which deals with the similar but not exactly the same task as ours. OFlow aims at performing shape completion for sequences of shapes while our DeepTracking-Net predicts the shape correspondence for consecutive frames. These two methods dealing with different tasks and the experimental settings are different. 

We evaluate and compare our proposed DeeoTracking-Net with Oflow\cite{niemeyer2019occupancy} on the D-Faust dataset. The D-Faust dataset is divided into 105 sequences for training sequences, 6 sequences for validation and 9 sequences for testing. 
Unlike OFlow which predicts the correspondence by encoding several observed time-varying shapes, our proposed DeepTracking-Net predicts the correspondence based on the previously observed motions. For the testing process, OFlow subdivides the testing sequence into small sub-sequence and reports the average performance. While, our model directly consumes the observed 3D shapes at consecutive three frames and approximates the tracking result in the fourth frame. We adopt a TCD with $D_z=128$ during the training and evaluation.

\begin{table}[!htp]
\centering
\begin{tabular}{cccccccc}
\hline\hline
\multicolumn{8}{c}{D-Faust} 
\\\hline
\multicolumn{4}{c|}{Method} &\multicolumn{2}{c|}{Chamfer} &\multicolumn{2}{c}{Correspond.}
\\\hline

\multicolumn{4}{l|}{PSGN 4D} &\multicolumn{2}{c|}{0.127} &\multicolumn{2}{c}{3.041}
\\

\multicolumn{4}{l|}{PSGN 4D (w/cor.)} &\multicolumn{2}{c|}{0.119} &\multicolumn{2}{c}{0.131}
\\

\multicolumn{4}{l|}{ONet 4D} &\multicolumn{2}{c|}{0.140} &\multicolumn{2}{c}{-}
\\

\multicolumn{4}{l|}{OFlow} &\multicolumn{2}{c|}{0.095} &\multicolumn{2}{c}{0.149}
\\

\multicolumn{4}{l|}{OFlow (w/cor.)} &\multicolumn{2}{c|}{0.084} &\multicolumn{2}{c}{0.117}
\\

\multicolumn{4}{l|}{\textbf{Ours}} &\multicolumn{2}{c|}{\textbf{0.004}} &\multicolumn{2}{c}{\textbf{0.035}}
\\


\hline\hline

\end{tabular}
\caption{
Results on the D-Faust dataset. We report  Chamfer distance and correspondence $\ell_2$-distance. }
\label{tab:dfaustsyn_quant}
\end{table}

\noindent\textbf{Result Analysis: } 
The quantitative results of tracking on consecutive frames can be found in the left part of Table \ref{tab:dfaustsyn_quant}. Our model achieves a Chamfer distance of $0.004$ and an average distance of $0.035$, while the OFlow (w/corr.) method achieves a Chamfer distance of $0.08$ and an average distance of $0.082$. Note that, our method is completely unsupervised while the OFlow (w/corr.) requires the full supervision of ground truth labels. We also note that the differences in methods and experimental settings might contribute to the fact that our method largely outperforms OFlow. 

\subsection{3D Tracking on Temporally Inconsistent Data}


\begin{figure}[ht!]
\centering
  \includegraphics[width=\textwidth,height=5.5cm]{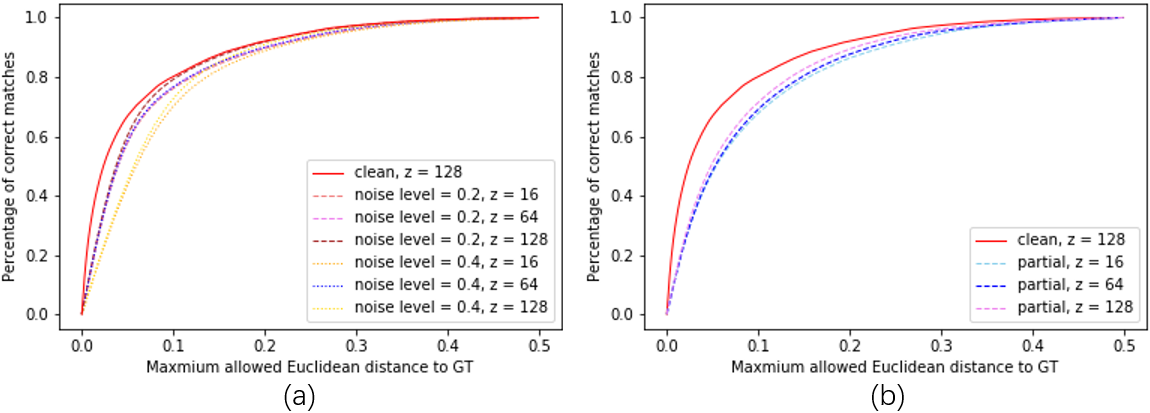}
  \caption{
  (a). The comparison of matching accuracy under different noise levels. (b). The matching accuracy on partial data of the SynMotion dataset.
  }
  \label{fig:inc_cur} 
\end{figure}
\noindent\textbf{Experiment Settings:} In this section, we conduct experiments that test our model's ability to handle data with noise and partial observations.
To construct the inconsistent data from SynMotions, we add Gaussian noise to each point of the shape with zero-mean and a standard derivation of 0.02 and 0.04 to create our SynMotions-noisy dataset. Also, we randomly sample 300 points from each shape to form our SynMotions-partial dataset. We set the dimension of our TCD to be 128 in our experiments. We directly use the trained model in section \ref{sc_clean} and optimize TCDs with fixed network parameters $\phi$ and temporal weight $\omega$. 

\noindent\textbf{Result Analysis: } 
Table \ref{tab:syn_quant} listed the quantitative results which show the promising performance of our model to handle inconsistent data with noise and partial observations. As we can see in this table, our model only experiences a slight performance drop with inconsistent inputs. Moreover, from the matching accuracy curve in Figure \ref{fig:inc_cur}, we can find that our DeepTracking-Net demonstrates a good 3D tracking performance with a slight performance drop, which further verifies its ability to predict a reasonable tracking result against inconsistent inputs. However, the performance drops when the shapes encounter with large noise level.

\begin{table}[!htp]
\centering
\begin{tabular}{cccccccc}
\hline\hline
\multicolumn{8}{c}{SynMotion} 
\\\hline
\multicolumn{4}{c|}{Method} &\multicolumn{2}{c|}{Chamfer} &\multicolumn{2}{c}{Correspond.}
\\\hline

\multicolumn{4}{l|}{Ours, Clean Data} &\multicolumn{2}{c|}{0.016} &\multicolumn{2}{c}{0.071} 
\\

\multicolumn{4}{l|}{Ours, Noise Level=0.2} &\multicolumn{2}{c|}{0.016} &\multicolumn{2}{c}{0.074}
\\

\multicolumn{4}{l|}{Ours, Noise Level=0.4} &\multicolumn{2}{c|}{0.015} &\multicolumn{2}{c}{0.080}
\\

\multicolumn{4}{l|}{Ours, Partial} &\multicolumn{2}{c|}{0.022} &\multicolumn{2}{c}{0.075} 
\\

\hline\hline

\end{tabular}
\caption{
Results on the SynMotion dataset. We report  Chamfer distance and correspondence $\ell_2$-distance. }
\label{tab:syn_quant}
\end{table}

\subsection{Temporal Correspondence Descriptor vs. Non-temporal Correspondence Descriptor}

\noindent\textbf{Experiment Settings:} 
In this experiment, we evaluate our design of using temporal-aware descriptor $Z$ instead of non-temporal descriptor $S$. We conduct further experiments on SynMotions clean dataset using the geometric content of each frame, but without connections among frames. That is, we directly concatenate $s_i$ of each frame to its corresponding state for training and testing.

\noindent\textbf{Result Analysis: }

\begin{table}[!htp]
\centering
\begin{tabular}{cccccccc}
\hline\hline
\multicolumn{8}{c}{SynMotions} 
\\\hline
\multicolumn{4}{c|}{Method} &\multicolumn{2}{c|}{Chamfer} &\multicolumn{2}{c}{Correspond.}
\\\hline

\multicolumn{4}{l|}{Ours} &\multicolumn{2}{c|}{0.016} &\multicolumn{2}{c}{0.071} 
\\

\multicolumn{4}{l|}{Ours w/o TSLC} &\multicolumn{2}{c|}{0.031} &\multicolumn{2}{c}{0.102}  
\\

\hline\hline
\end{tabular}
\label{tab:abl}
\caption{
Ablation study results on clean SynMotions clean dataset. The table shows the results of our model with and without temporal information. We report Chamfer distance (lower is better) and correspond $\ell_2$-distance (lower is better). 
}
\end{table}

The quantitative results are shown in the right part of Table 4. From the table, one can see that our method with the TCD performs a lot better than directly concatenate $s_i$ of each frame to its corresponding state. This demonstrates the effectiveness of using a temporal-aware correspondence descriptor instead of a non-temporal-aware one.

\section{CONCLUSIONS}
In this paper, we proposed an unsupervised method for time-varying 3D tracking. To avoid the challenges from the explicit design of the feature encoding network, we designed a novel learnable latent vector to encode both temporal and spatial essence for time-varying shape correspondence for 3D tracking. Based on this temporal-aware correspondence descriptor, we design a continuous flow decoder network to produce the displacement from one frame to the next. Moreover, we introduce a new benchmark dataset SynMotions for the 3D tracking task that contains dense ground-truth correspondences between consecutive frames. Experiments on SynMotions and D-FAUST datasets prove the effectiveness of our proposed method for the task of 3D tracking and our proposed method can produce promising tracking results and it even outperforms the supervised approach.

\clearpage
%
%
\bibliographystyle{splncs04}
\bibliography{main}
\end{document}


\title{Supplementary Material for \\DeepTracking-Net: 3D Tracking with Unsupervised Learning of Continuous Flow} 

\titlerunning{DeepTracking-Net}
%
\author{Shuaihang Yuan
\and
Xiang Li
\and
Yi Fang\thanks{indicates corresponding author.}
}
%
\authorrunning{S. Yuan, X. Li and Y. Fang}
%

\institute{NYU Multimedia and Visual Computing Lab, USA \and
New York University Abu Dhabi, UAE\and
New York University, USA\\
\email{\{sy2366, xl1845, yfang\}@nyu.edu}}
\maketitle              

\begin{abstract}
In this supplementary material, we first give experimental details in Section 1. Then, we state the reason why we compare with our method with Occupancy Flow \cite{niemeyer2019occupancy}, the differences between two methods and why we outperform Occupancy Flow \cite{niemeyer2019occupancy} by a great margin in Section 2. Finally, in Section 3, we illustrate how we generate the proposed SynMotion dataset.
\keywords{3D tracking, Deep neural networks, Spatiotemporal embedding, Correspondence}
\end{abstract}

\section{Experimental settings}
In this section, we go through the details of how we conduct our experiment. We adopt two datasets for training and testing, including Dynamic FAUST \cite{dfaust:CVPR:2017} dataset and our proposed SynMotion. For all the experiments, we use Adam optimizer \cite{kingma2014adam} with a learning rate of 0.001 and batch size is set to 32. During training, our model is trained for 1000 iterations; while in the testing stage, we optimize our temporal correspondence descriptors (TCDs) for 1000 iterations. For all experiments, we first group the dataset to a sequence of four frames, i.e., the input at time stamp $i$ can be expressed $X_i=(M_{i-2},M_{i-1},M_{i}, M_{i+1})$, where $M_{i-2},M_{i-1},M_{i}$ are three previously observed frames and $M_{i}$ is the current unseen frame which we use to evaluate the tracking performance. For each shape in a training sample, we sample 2000 points uniformly to formulate our input and compute unsupervised loss $\mathcal{L}_{\phi,\mathcal{Z},\omega}$ as a supervision for the optimization of CF-Decoder network as well as the TCDs. In the inference stage, with the optimized CF-Decoder network and TCDs, our model can directly predict temporally and spatially continuous flow fields from shape $M_{i}$ to $M_{i+1}$,  and we mark the predicted shape as $\mathcal{T}_{M_i}$. Note that there exists an identical correspondence between $M_i$ and $\mathcal{T}_{M_i}$. To find correspondences between shape $M_{i}$ and shape $M_{i+1}$, we use a nearest neighbor search to find correspondences between shape $M_{i+1}$ and the predicted shape $\mathcal{T}_{M_i}$.


\section{Comparison}
In this section, we first illustrate the reason why we compare our method with Occupancy Flow (OFlow) \cite{niemeyer2019occupancy} and then we state the differences between these two methods. Finally, we give the reasons why our method outperforms Occupancy Flow by a large margin.

\subsection{Motivation}
One reason motivates us to compare our proposed method with the OFlow is that our propose method aims at solving the 3D tracking problem by considering both spatial and temporal correlation among shapes. The OFlow, to the best of our knowledge, is the newest and only state-of-art approach that also leverage the spatial and temporal information to solve 3D geometric problem, which is highly related to our proposed method. The OFlow and our method both compute a continuous field in the spatial and temporal domains which can implicitly yield correspondences among the consecutive 3D frames. The performance of OFlow and our DeepTracking-Net are both evaluated based on the quality of dense correspondences. Therefore, we compared our method with OFlow in our experiments in terms of the quality of estiamted dense corrspodnences.


In our work, we propose the temporal correspondence descriptor (TCD) that describe the transformation relationships between the each state to ensure the temporal continuity the and the continuous flow decoder to decode a continuous field from TCD and current frame which, similar to OFlow, not only yield implicit correspondences but also preserve continuities in spatial and temporal domains.

\subsection{Differences}

Our model is different from OFlow from two perspectives. First, the OFlow model takes all frames as input and use a neural network to explicit learn the global temporal feature representations; while in our DeepTracking-Net, the temporal relationships are explicitly embedded by propagating the TCD of the previous frame to the next frame in order to capture the locally integrated features (i.e. feature of each frame). 

Second, the OFlow aims at performing shape completion for sequences of shapes, the shape correspondences are implicitly considered among all input shapes as a whole with an intention to get spatio-temporal consistent completion. In our DeepTracking-Net, the correspondence prediction for each frame are calculated from several preceding frames only---independent of other non-neighboring frames, the computation is coupled across all frames, thus makes our method able to generate globally consistent correspondence by piecing together all local information.



\begin{table}[!htp]
\centering
\begin{tabular}{cccccccc}
\hline\hline
\multicolumn{8}{c}{D-Faust}  
\\\hline
\multicolumn{4}{c|}{Method} &\multicolumn{2}{c|}{Chamfer} &\multicolumn{2}{c}{Correspond.}
\\\hline

\multicolumn{4}{l|}{PSGN 4D} &\multicolumn{2}{c|}{0.127} &\multicolumn{2}{c}{3.041} \\

\multicolumn{4}{l|}{PSGN 4D (w/cor.)} &\multicolumn{2}{c|}{0.119} &\multicolumn{2}{c}{0.131} \\

\multicolumn{4}{l|}{ONet 4D} &\multicolumn{2}{c|}{0.140} &\multicolumn{2}{c}{-} \\

\multicolumn{4}{l|}{OFlow} &\multicolumn{2}{c|}{0.095} &\multicolumn{2}{c}{0.149} \\

\multicolumn{4}{l|}{OFlow (w/cor.)} &\multicolumn{2}{c|}{0.084} &\multicolumn{2}{c}{0.117} \\

\multicolumn{4}{l|}{\textbf{Ours}} &\multicolumn{2}{c|}{\textbf{0.004}} &\multicolumn{2}{c}{\textbf{0.035}} \\

\hline\hline

\end{tabular}
\caption{
Quantitative results on D-Faust dataset. Chamfer distance and correspondence $\ell_2$-distance are reported. }
\label{tab:dfaust_quant}
\end{table}

\subsection{Result Analysis}
Our model and OFlow are evaluated based on the same evaluation metrics as OFlow, including the correspondences distance and the Chamfer distance, presented in Table \ref{tab:dfaust_quant}. The reason why our proposed model outperforms the OFlow is mainly because our model predicts shape correspondence from preceding frames in a \textbf{local window}, which makes it pay much more attention to its neighbor frames; while OFlow performs shape completion for all inputs frames simultaneously and the prediction results for each frame are generated with a \textbf{global consistency constrain}. Especially, in OFlow, they infer a sequence of 3D geometries from the reference 3D shape at $T=0$ which accumulate the error when $T$ increases. In the way we proposed in our experiments, we can make sure that our proposed method gets rid of the error accumulation.

\section{SynMotion Dataset}
\begin{figure}[ht!]
\centering
  \includegraphics[width=\linewidth]{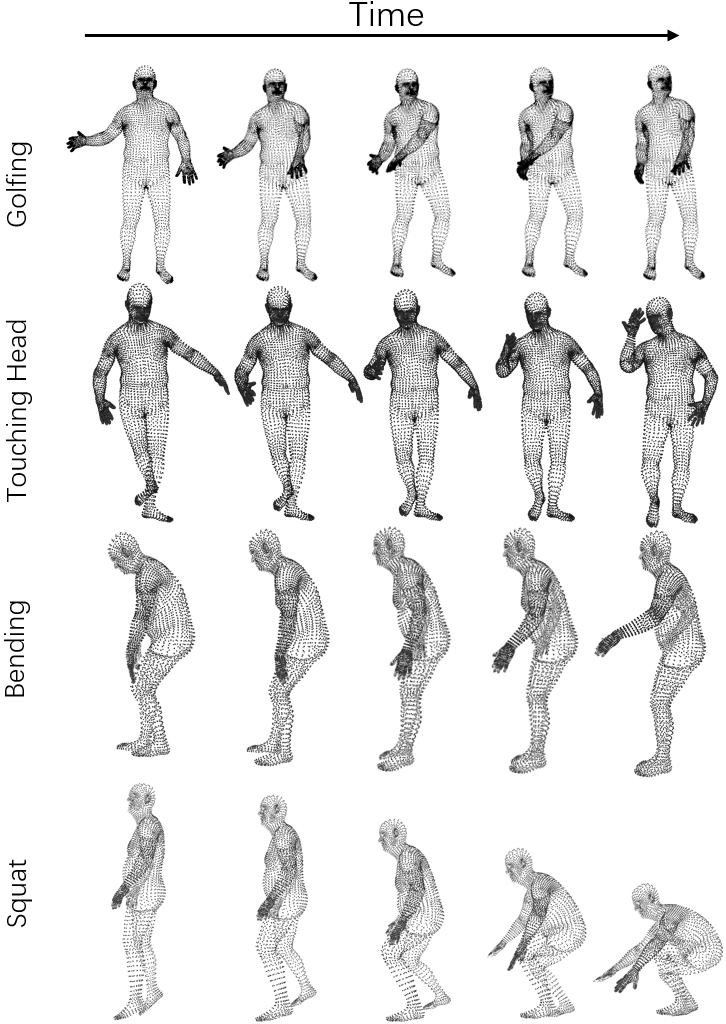}
  \caption{
    The visualization samples of the SynMotion dataset.
  }
  \label{fig:dataset_sup} 
\end{figure}

In this section, we present how we generate the SynMotion dataset.
To generate a sequence of time-varying motion dataset, we first adopt the SMPL human model \cite{SMPL:2015} as template models and the pose of SMPL model is governed by 72 parameters. The detail of SMPL human model can be found in Section 4.2 of the main paper. Since there are no predefined limitations for those parameters, the pose can be distorted if we randomly assign values to those parameters. To obtain reasonable human poses, we adopt the SURREAL pose database \cite{varol17_surreal}. Once we obtain human poses, we randomly select two poses and we interpolate those 72 parameters long the time dimension between the selected two poses. Although the deformation levels between two randomly select poses are different, we interpolate those parameters for a certain number of frames to simulate the speeding-up or slowing-down process in a sequence of motions. The visualization of our proposed dataset can be found in Figure \ref{fig:dataset} and Figure \ref{fig:dataset_sup}

\begin{figure}[ht!]
\centering
  \includegraphics[width=\linewidth]{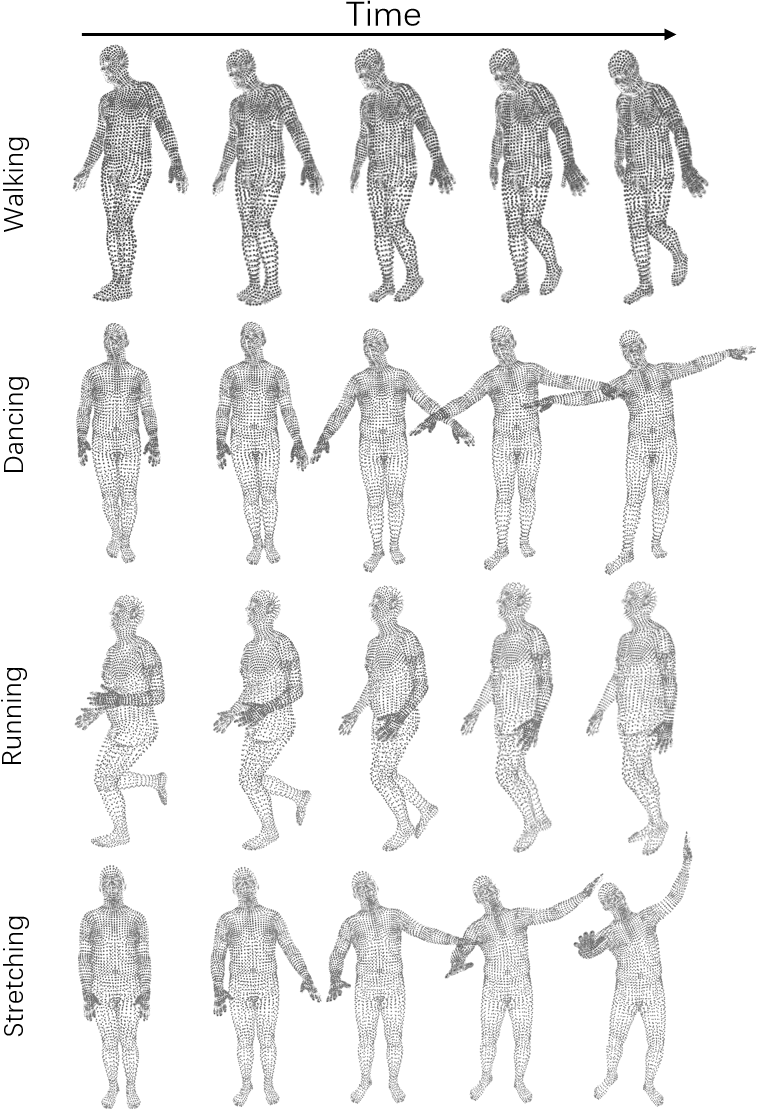}
  \caption{
    The visualization samples of the SynMotion dataset.
  }
  \label{fig:dataset} 
\end{figure}

\begin{figure}[ht!]
\centering
  \includegraphics[width=\linewidth]{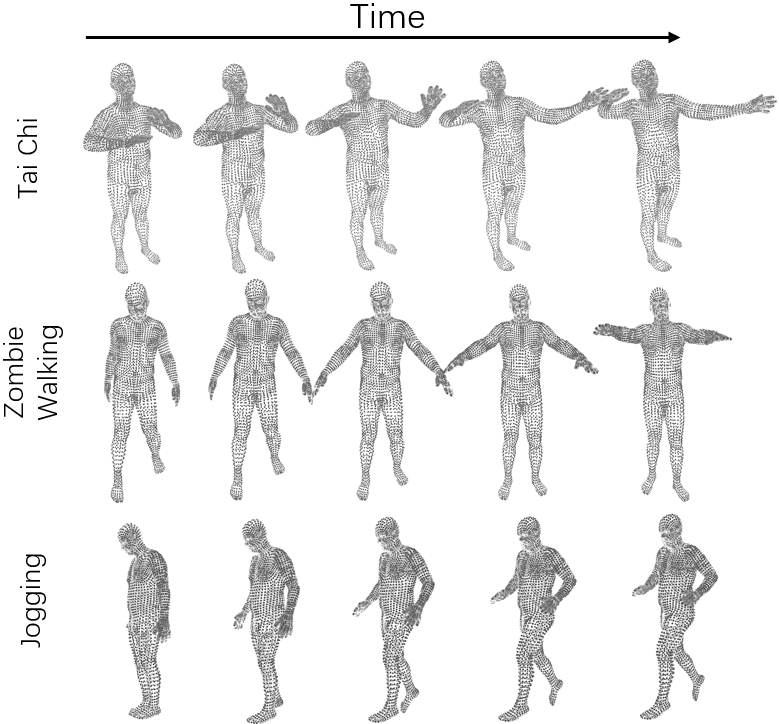}
  \caption{
    The visualization samples of the SynMotion dataset.
  }
  \label{fig:dataset} 
\end{figure}

\clearpage
%
%
\bibliographystyle{splncs04}
\bibliography{egbib}